\documentclass{article}

\usepackage{PRIMEarxiv}

\usepackage[utf8]{inputenc} 
\usepackage[T1]{fontenc}    
\usepackage{hyperref}       
\usepackage{url}            
\usepackage{booktabs}       
\usepackage{amsfonts}       
\usepackage{nicefrac}       
\usepackage{microtype}      
\usepackage{lipsum}
\usepackage{fancyhdr}       
\usepackage{graphicx}       
\graphicspath{{media/}}     
\usepackage{multicol}
\usepackage{graphicx}
\usepackage{setspace}
\usepackage{hyphenat}
\usepackage[numbers]{natbib}
\usepackage{amsmath}
\usepackage{algorithm}
\usepackage{algpseudocode}
\usepackage{mathtools}
\newcommand{\norm}[1]{\left\lVert#1\right\rVert}

\title{Average Outward Flux Skeletons for Environment Mapping and Topology Matching
	\thanks{\textit{\underline{Disclaimer}}: 
		Babak Samari and Gregory Dudek contributed to this article in their personal capacity as Ph.D. student and Professor at McGill University, respectively.}
}

\begin{document}
	\maketitle
	\vspace{-2cm}
	\begin{center}
		\textbf{\large Morteza Rezanejad$^{123}$, Babak Samari$^{4}$, Elham Karimi$^{5}$,\\[0.3cm] Ioannis Rekleitis$^{6}$, Gregory Dudek$^{4}$, Kaleem Siddiqi$^{4}$}\\[.5cm]
		\textit{$^{1}$Department of Psychology, University of Toronto, Toronto, Canada}\\
		\textit{$^{2}$Department of Computer Science, University of Toronto, Toronto, Canada}\\
		\textit{$^{3}$Department of Mechanical \& Industrial Engineering, University of Toronto, Toronto, Canada}\\
		\textit{$^{4}$School of Computer Science and Centre for Intelligent Machines, McGill University, Montréal, Canada}\\
		\textit{$^{5}$Rosalind and Morris Goodman Cancer Research Centre, McGill University, , Montréal, Canada}\\
		\textit{$^{6}$Computer Science and Engineering Department, University of South Carolina, Columbia, United States}\\[1cm]
	\end{center}

	\begin{abstract}
		We consider how to directly extract a road map (also known as a topological representation) of an initially-unknown 2-dimensional environment via an online procedure that robustly computes a retraction of its boundaries.  While such approaches are well known for their theoretical elegance, computing such representations in practice is complicated when the data is sparse and noisy. In this article, we first present the online construction of a topological map and the implementation of a control law for guiding the robot to the nearest unexplored area, first presented in \cite{rezanejadrobust}. The proposed method operates by allowing the robot to localize itself on a partially constructed map, calculate a path to unexplored parts of the environment (frontiers),  compute a robust terminating condition when the robot has fully explored the environment, and achieve loop closure detection. The proposed algorithm results in smooth safe paths for the robot's navigation needs. The presented approach is an any time algorithm that has the advantage that it allows for the active creation of topological maps from laser scan data, as it is being acquired. The resulting map is stable under variations to noise and the initial conditions. We also propose a navigation strategy based on a heuristic where the robot is directed towards nodes in the topological map that open to empty space. The method is evaluated on both synthetic data and in the context of active exploration using a Turtlebot 2. Our results demonstrate a complete mapping of different environments with smooth topological abstraction without spurious edges.  We then extend the work in \cite{rezanejadrobust} by presenting a topology matching algorithm that leverages the strengths of a particular spectral correspondence method, FOCUSR \cite{lombaert2012focusr}, to match the mapped environments generated from our topology making algorithm. Here, we concentrated on implementing a system that could be used to match the topologies of the mapped environment by using AOF Skeletons. In topology matching between two given maps and their AOF skeletons, we first find correspondences between points on the AOF skeletons of two different environments. We then align the (2D) points of the environments themselves.
		We also compute a distance measure between two given environments, based on their extracted AOF skeletons and their topology, as the sum of the matching errors between corresponding points. We evaluate our topology matching algorithm and demonstrate promising results on a few environments of increasing complexity, with simulated sensor noise.
	\end{abstract}
	
	\section{Introduction}
	
	Our approach is based on AOF skeletons from two-dimensional, dense, laser data.  This fundamentally one-dimensional structure (embedded in 2D)
	constitutes a robust, efficient elegant representation that can be used for several applications \cite{rezanejad2015view,rezanejad2019gestalt} including a range of navigation and localization tasks.
	
	Topological representations have been proposed and employed in robotics for over 25 years~\cite{kuipers1991robot,chatila1985position,dudek1991robotic} because of the potentially simple ensuing control laws,  their relevance to human cognitive mapping, and their simplicity.   At the core of many approaches to extracting topological representations from real environments is the calculation of points that are locally maximally distant from, sensed obstacles, yielding the medial axis. Such topological structures result in safe areas where robots can navigate without collisions. Indoor environments with long corridors, like those found in malls and office buildings, and underground mines, are ideal candidates for using a topological representation.
	
	\begin{figure}[!htb]
		\centering
		\includegraphics[width=0.9\textwidth]{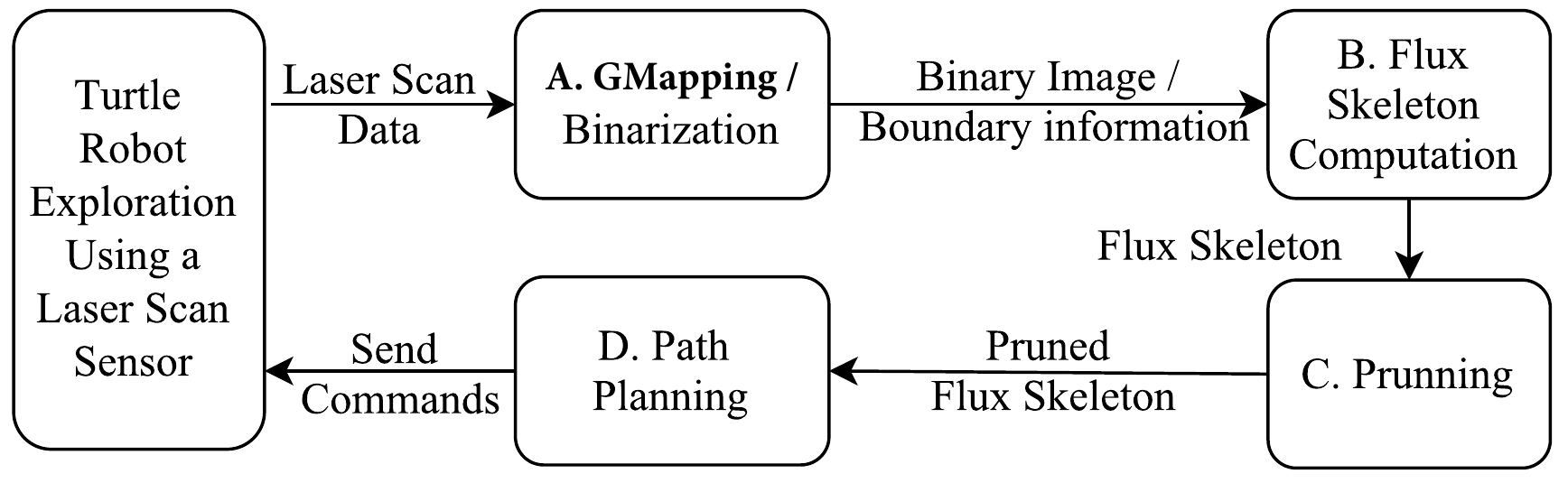}
		\caption{
			System overview. The system consists of four independently running modules along with a robot which is exploring the environment. Each of these modules is a component of a feedback chain system.} 
		\label{fig:systemflowchart}
	\end{figure}

\begin{figure}[!t]
		\centering
		\begin{tabular}{c@{\hskip 2mm}c@{\hskip 1mm}c@{\hskip 1mm}c}
			\includegraphics[height=0.25\textwidth]{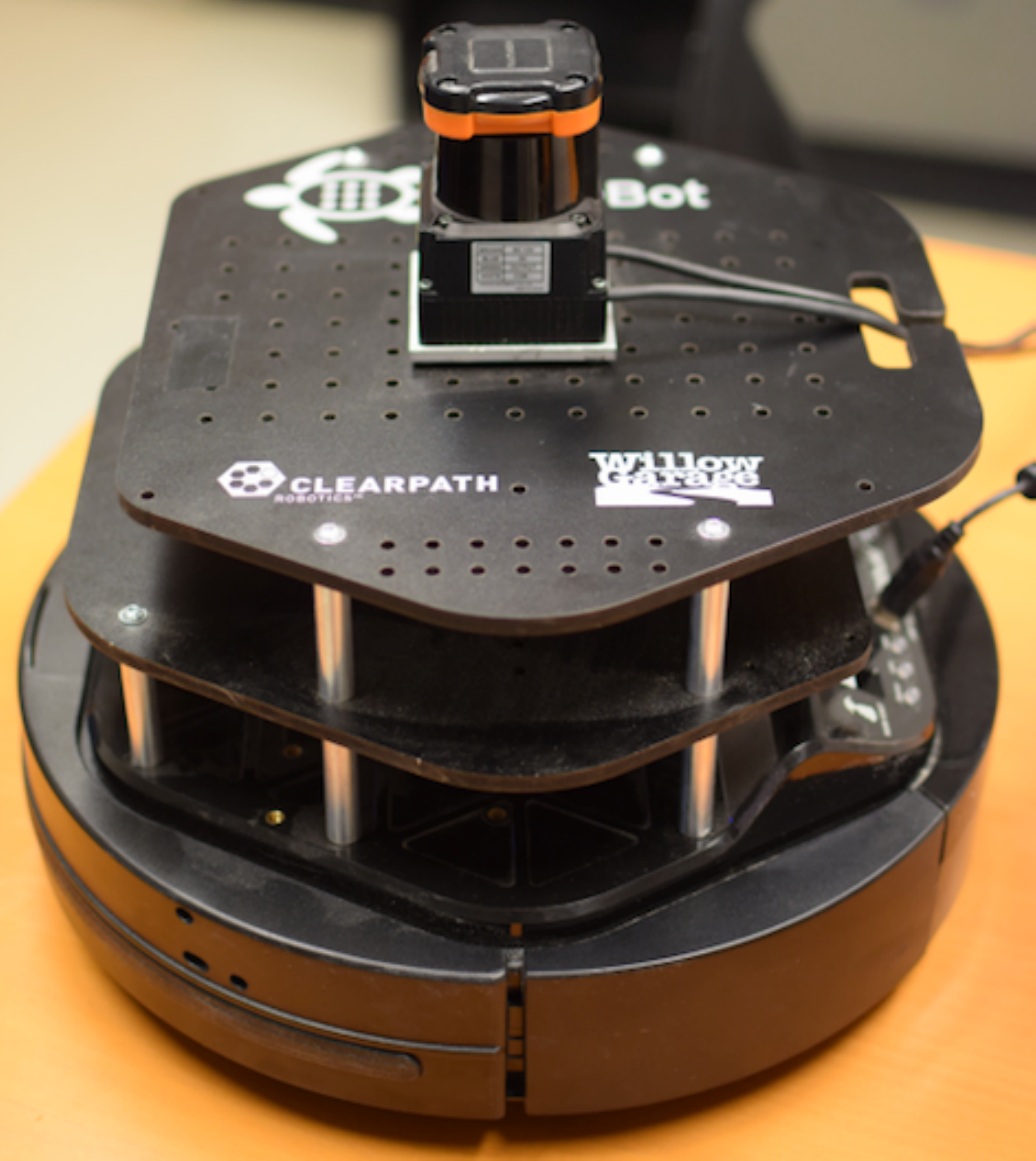}  &  \includegraphics[height=0.25\textwidth]{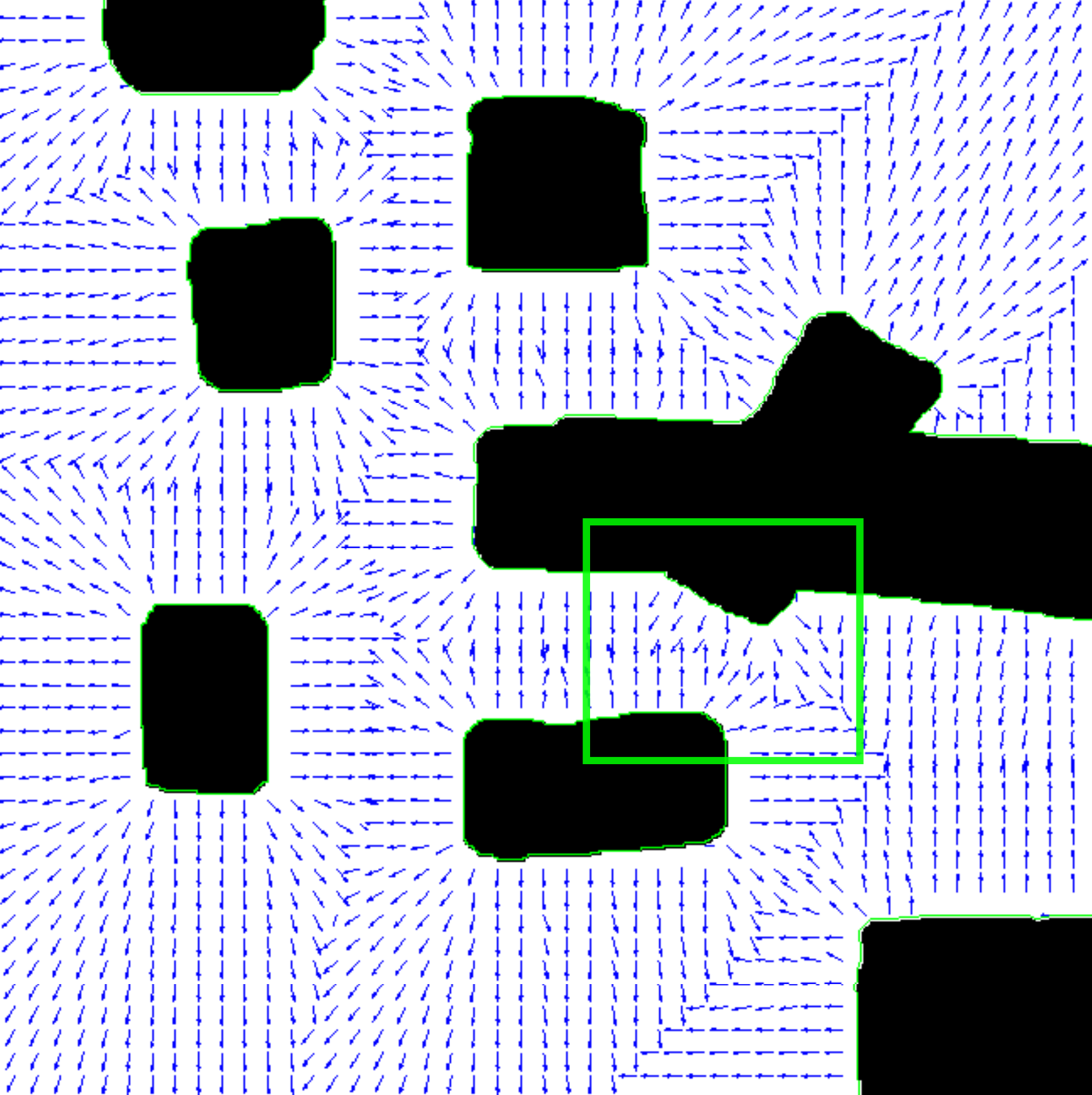} & \includegraphics[height=0.25\textwidth]{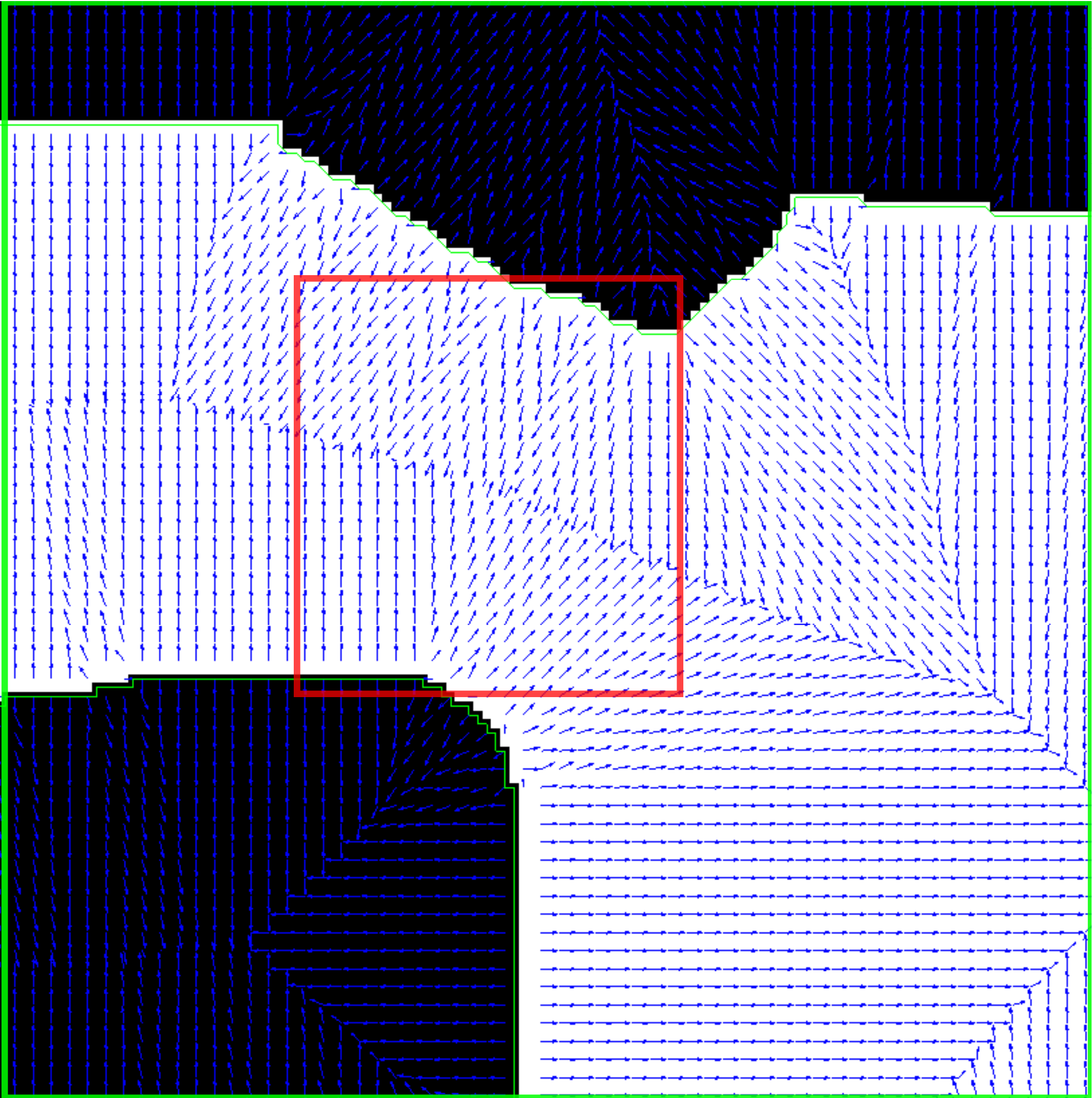}& \includegraphics[height=0.25\textwidth]{Figures/GVF-Fig1}\\
			\textbf{(a)} & \multicolumn{3}{c}{\textbf{(b)}} 
		\end{tabular}
		\caption{\textbf{(a)}: the experimental platform used, a Turtlebot 2, with a Hokuyo laser range finder. \textbf{(b)}: an illustration of the Euclidean distance function gradient vector field $\dot{\mathbf{q}}$ for a sample environment where the black regions represent obstacles. Computation of AOF which is based on the integral of the Euclidean distance function gradient vector field $\dot{\mathbf{q}}$ provides a stable skeleton computation.}
		\label{fig:turtlebot_gradientVectorField}
	\end{figure}

	The major weakness of traditional skeletonization algorithms is that they suffer from high sensitivity to noise (perturbations of the boundary data).  A small perturbation in the sensed data can drastically change the skeleton structure and its abstraction.  This has led to several different results on the stability of the medial axis, including approaches that try to remove those skeletal branches that are likely to be generated due to boundary \cite{Chazal2004-ek,Attali2009-pn,Chazal2005-va,camaro2020appearance,rezanejad2020medial}.  In the present work, we opt for AOF skeletons because they yield a rather direct and robust manner for finding skeletal points since the computation of AOF involves integrals rather than derivatives (see Figure \ref{fig:turtlebot_gradientVectorField} \textbf{(b)}). Here skeletal points are associated with locations where the average outward flux of the gradient of the Euclidean distance function through a shrinking circular neighborhood is non-zero. We shall later deal with changes to the topological structure of the skeleton due to boundary noise not by altering the representation itself, but by using a matching algorithm that employs spectral signatures.
	
	The appeal of topological representations for mapping, exploration, and human-robot interaction has been noted by several authors who suggested they be used directly, computed from 2D data~\cite{kuipers1991robot,chatila1985position}.  The Voronoi diagram, a classic structure in computational geometry that has appeared in many fields, and the Generalized Voronoi Graph (GVG) have been exploited in robotics as a mechanism for computing topological maps~\cite{Choset1995,choset1997incremental}. Pure topological results have been studied in \cite{Dudek1989d,dudek1991robotic,Dudek93} for mapping a graph\hyp like world with minimal sensor input. More recent work includes~\cite{Wang2008} on exploration strategies on a graph\hyp like world. 
	
	The full employment of the GVG in a SLAM framework was proposed in~\cite{Choset01}, and extended for use in a hybrid metric/topological maps in~\cite{Thrun1998,Rekleitis2005}. Tully et al. \cite{Tully09} recommend a hypothesis tree method for loop\hyp closure where the branches that are considered to be
	unlikely based on topological and metric  GVG information are pruned. Among the pruning tests, it is worth mentioning the planarity test which ensures that when a loop\hyp closure has been decided, the resulting GVG graph remains planar. The utility of this test has been examined extensively in~\cite{Savelli04}. The purely topological variant of these approaches had been previously examined in \cite{Dudek93}.
	
	\cite{kuipers2004local} recommend the use of a framework, termed a hybrid spatial semantic hierarchy, where the incremental construction of topological large\hyp scale maps is employed in conjunction with metric SLAM methods for the creation of maps of small\hyp scale. No use of a global frame of reference is made and a multi\hyp hypothesis approach is used to represent potential loop\hyp closures.
	
	\section{Mapping Environments using Average Outward Flux Skeletons}
	\label{sec:map}    
	Our system takes laser scanned data from a 2D laser line scanner and generates an abstraction of the scanned environment. This is done through a number of modules: GMapping/binarization, average outward flux skeleton computation, pruning and simplification, and path planning for further exploration; see Figure  \ref{fig:systemflowchart}.  These modules are executed in a serial pipeline where the output of each module is the input to the next module.

	\subsection{GMapping and Binarization}
	\label{GMappingSubsection}
	
\begin{figure}[!htb]
		\centering
		\begin{tabular}{c@{\hskip20pt}c}
			\includegraphics[width=0.45\textwidth]{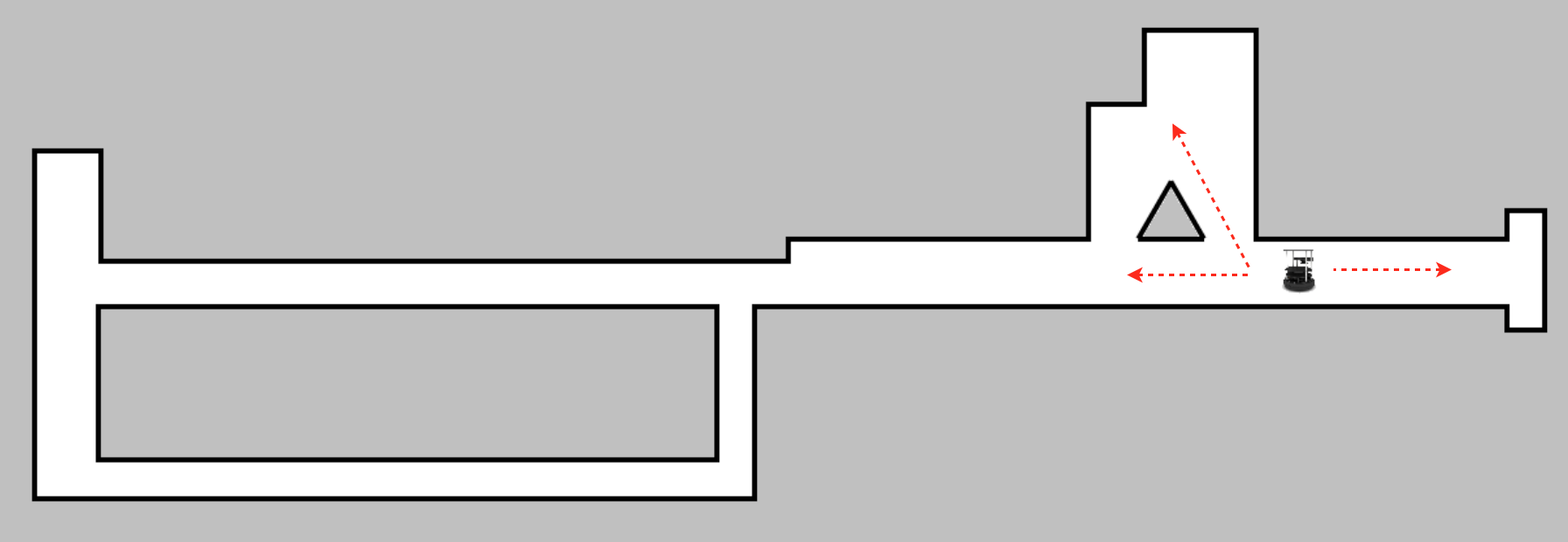} &
			\includegraphics[width=0.45\textwidth]{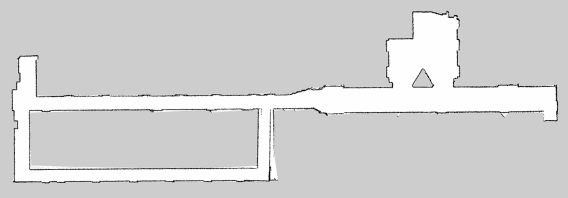}\\
			(a) & (b)\\
			\includegraphics[width=0.45\textwidth]{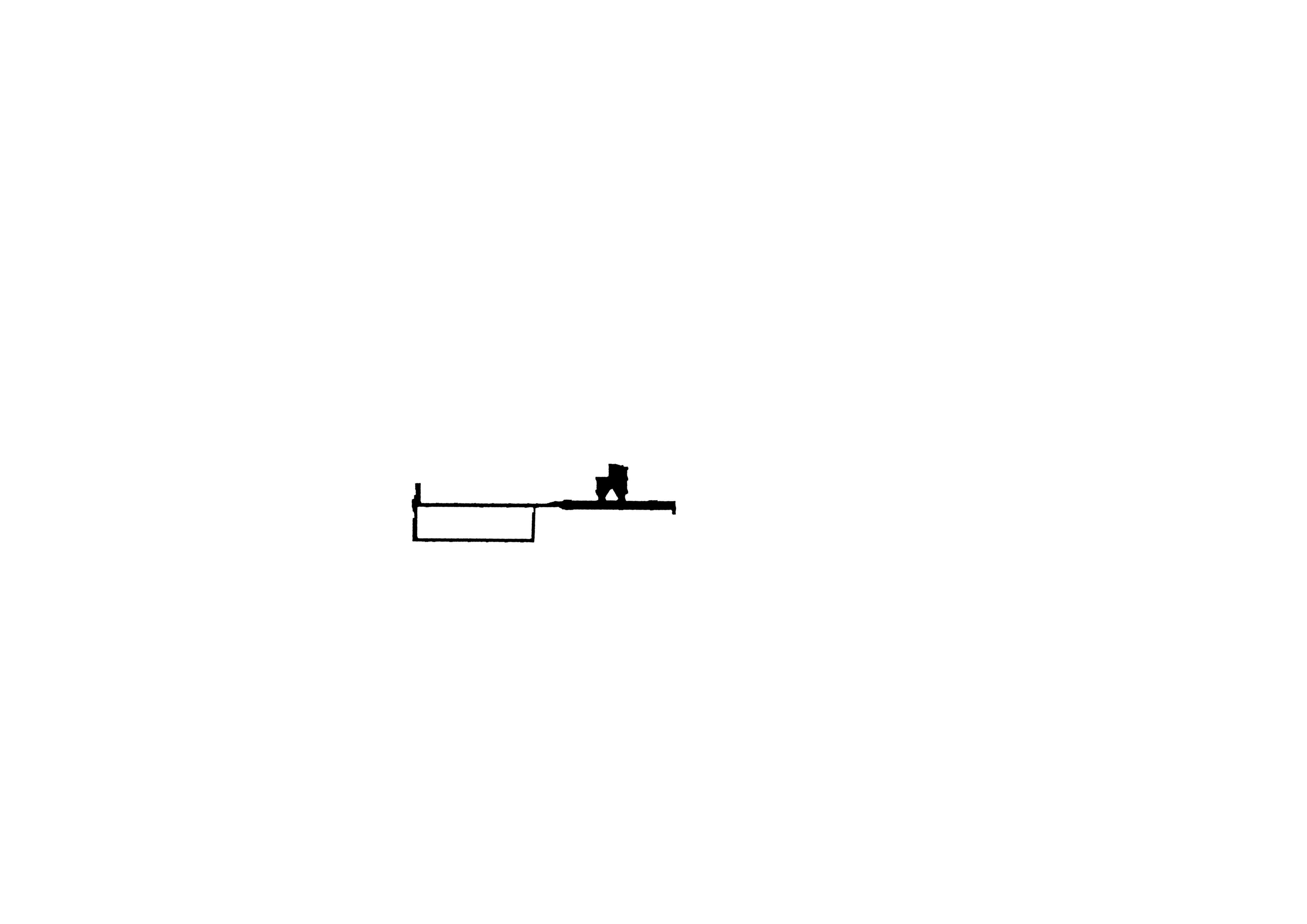} &
			\includegraphics[width=0.45\textwidth]{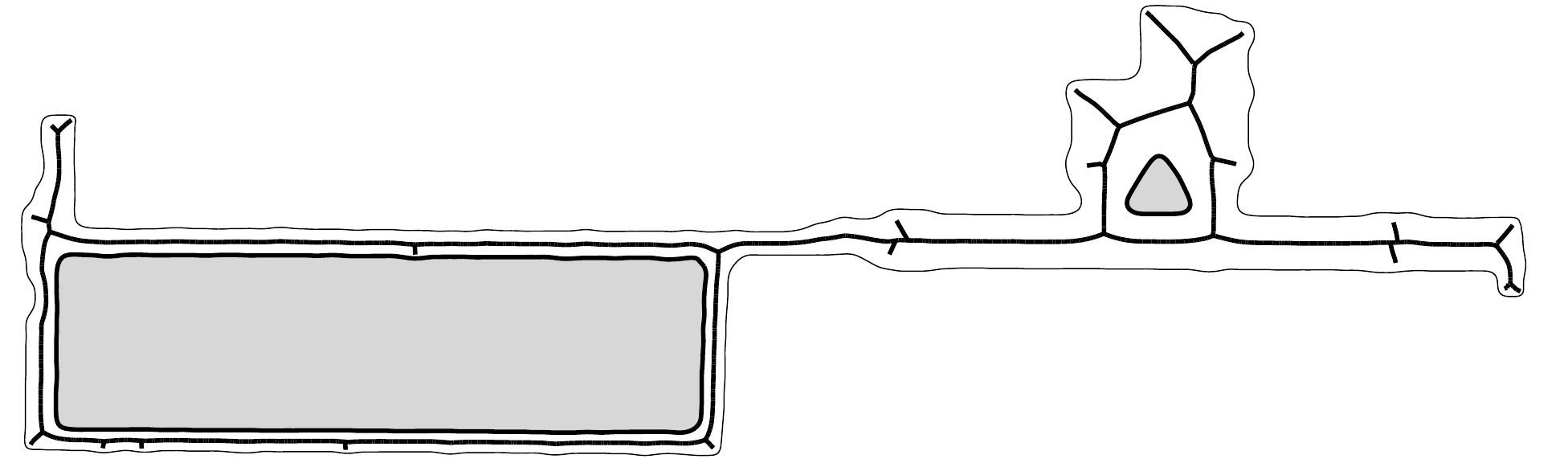}\\
			(c) & (d)\\
			\includegraphics[width=0.45\textwidth]{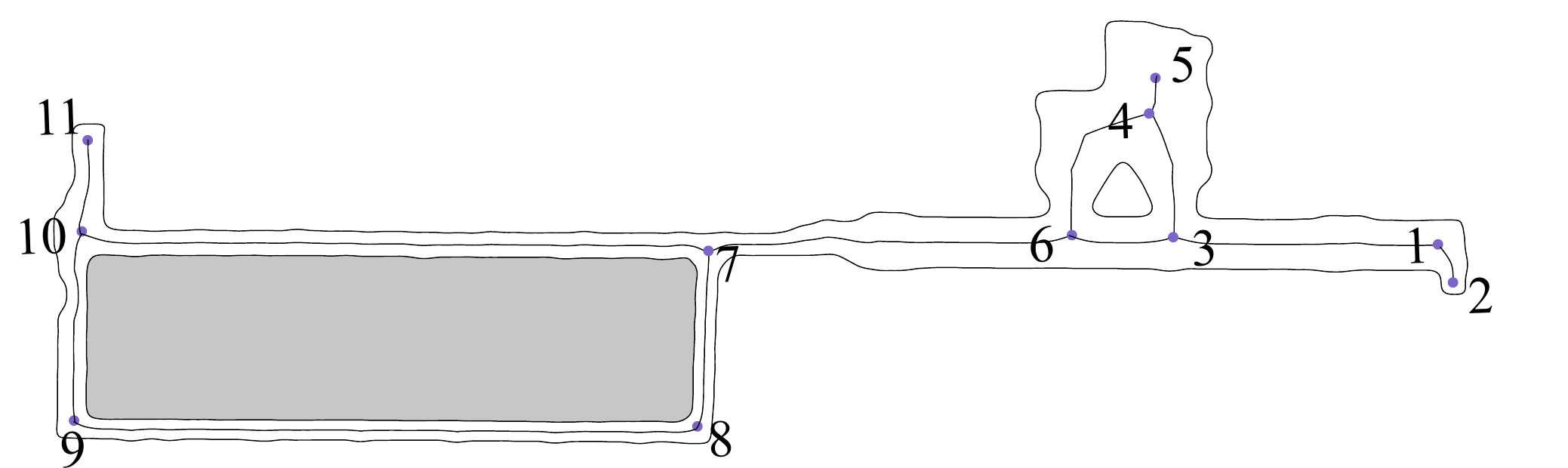} &
			\includegraphics[width=0.23\textwidth]{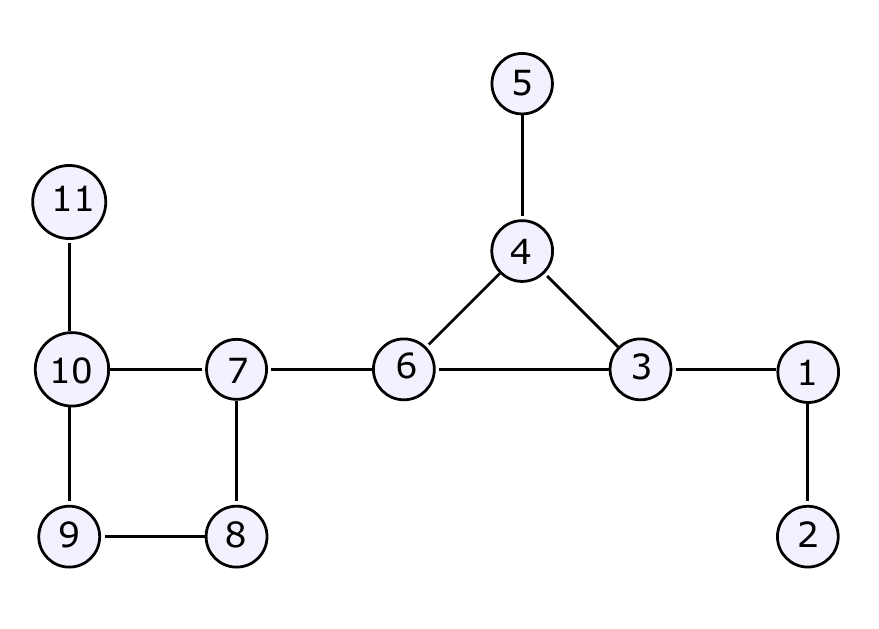}\\
			(e) & (f)\\
		\end{tabular}
		\caption{(a): An example map on an environment (b): The grid map of the environment. (c): the binarization of the grid map in the top row. (d): the full skeletonization process applied to the binarization of the environment. Although, the skeleton is very smooth, there are still branches that can be removed without altering its topology. (e): the skeleton in (d) is pruned and simplified in a way that makes robot navigation safe. Safe navigation means that robot does not go to an endpoint that is too close to a wall or an obstacle. (f): the topological map resulting from the abstraction in row four. Here, nodes are branch points or end points in the skeleton that are \textbf{not} removed by our pruning approach.}
		
		\label{fig:topolgy}
	\end{figure}

	The system first receives 2D laser scan data in a format where each scan is a single line containing range measurements. These laser scan data serve as input to the GMapping module.  GMapping is one of the most used laser-based SLAM algorithms \cite{grisetti2005improving}. It takes raw laser scan range data and odometry and produces gridmaps of the considered environment, where each gridmap is a probability distribution of cells (regions) being covered by the laser scan. The algorithm uses a highly efficient Rao-Blackwellized particle filter in which each particle has an individual map of the environment. The generated gridmap at the end of this stage is an intensity image where higher intensities show higher probabilities of being covered by the laser scanner (white regions), grey cells with lower intensities representing points that have not been covered yet by the robot, and where black cells usually represent walls where the range scanner has faced a physical obstacle.
	Figure  \ref{fig:topolgy}(b)  shows an example of a gridmap obtained after a certain amount of scanning.

	Gridmaps must be binarized before they can be fed as input to our average outward flux-based skeletonization algorithm. To do this we apply the following sequence of steps:    a) all pixels on gridmaps that are not scanned (gray regions) are set to background regions. Pixels that have a high probability of being obstacles (e.g. walls - black pixels) are stored as the foreground regions. b) Gaussian blurring is applied to smooth the structure that remains. c) The resulting image is then thresholded to give a binary one. d) The contours of all foreground regions in this image are extracted and sorted according to their area;  regions having very small areas are considered outliers. During this process, we keep track of the transformation needed to translate the final output to world coordinates. The second row in Figure  \ref{fig:topolgy}  depicts a binarized version of the grid map in the top row.
	
	\subsection{Pruning the Skeleton}
	\label{simplificationSubSection}
	


	As illustrated in Figure  \ref{fig:topolgy} (third row), the skeletonization process yields some branches that can be pruned without altering the topology of the skeleton. To prune such branches with the goal of topological mapping, we suggest a fairly simple but effective algorithm where the robot explores unseen regions and avoids getting too close to obstacles. 
	Our algorithm works as follows: it looks for all branches that have one branch point connected to an endpoint. If that endpoint is surrounded by walls of obstacles then there is no possibility for further exploration in that direction. In such a case, the branch connecting the branch point to the endpoint is removed and the skeleton is simplified.

	\section{Path Planning}
	\label{pathPlanningSubsection}
	\begin{figure}[!b]
		\centering
		\includegraphics[width=0.79\textwidth]{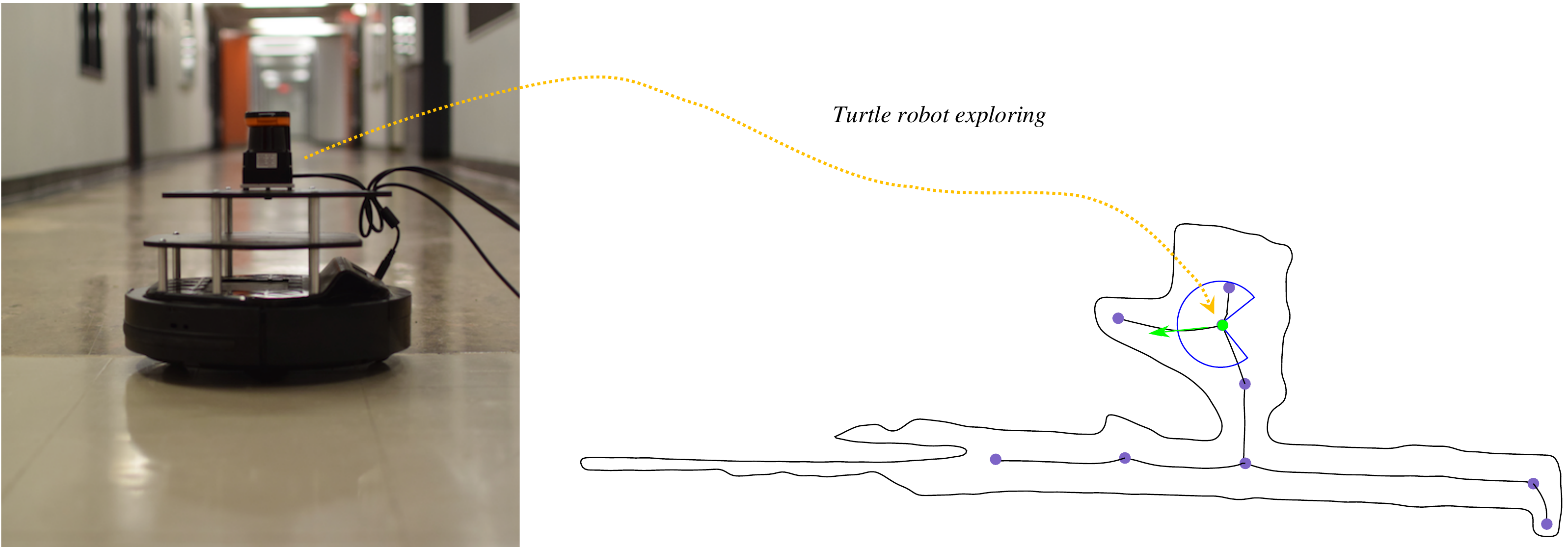}
		\caption{The exploring robot situated at one of the corridors of the McConnell Engineering Building at McGill University. This image is taken at the junction next to the triangular obstacle in the center of the map; see Figure  \ref{fig:topolgy}. The environment has been partially explored and the robot now selects an edge (green) leading into unexplored space. The Pacman shape represents the current position and direction of the heading of the exploring robot.}
		\label{fig:planning}
	\end{figure}
	The last step of this process is to guide the robot through the environment to explore new territory. As we showed in subsection \ref{simplificationSubSection}, at each time, there is a partial abstraction of the environment generated by utilizing the GMapping package, discretizing the output map of GMapping, and then extracting the AOF skeleton. These steps produce an up-to-date topological map of the environment at each time step. The system keeps track of the visited nodes in a list. This enables the system to explore novel territory for its next move. As mentioned before, at each time step the robot moves from one of the visited nodes to a frontier node; the new nodes traversed on\hyp route to the frontier node are added to the list of visited nodes. Each connecting edge is weighted by the length of the path through the skeleton. This weighting strategy results in a selection of good candidates for future exploration. In the algorithm, the nearest frontier node to the current node is selected. To find the nearest node, we use the \textit{Bellman Ford} algorithm \cite{bellman1958routing} which computes the shortest paths from a single node to all of the other nodes in a weighted directed graph; see Figure  \ref{fig:planning}.

	\section{Environment Mapping Experiments}
	\label{sec:results}
	
	Several experiments were performed, both in simulation and with a real robot. The proposed methodology was implemented under the ROS framework\footnote{http://www.ros.org/}. 
	
	\subsection{Experiments with a Real Robot}
	During the non-simulation experiments the Turtlebot 2 platform was used with a Hokuyo laser range finder; see Figure  \ref{fig:planning}. The laser sensor has a range of 30 meters and has a  $270^\circ$ field of view, returning a dense cloud of 1080 coplanar points. 
	
\begin{figure}[!tb]
		\centering
		\begin{tabular}{@{\hskip 0mm}c@{\hskip 4mm}c}
			\includegraphics[width=0.06\textwidth]{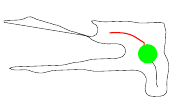} & \includegraphics[width=0.24\textwidth]{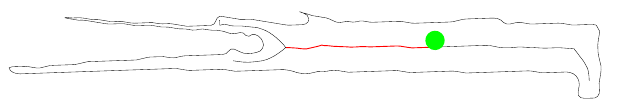} \\[1mm]
			(a) & (b) \\
			\includegraphics[width=0.24\textwidth]{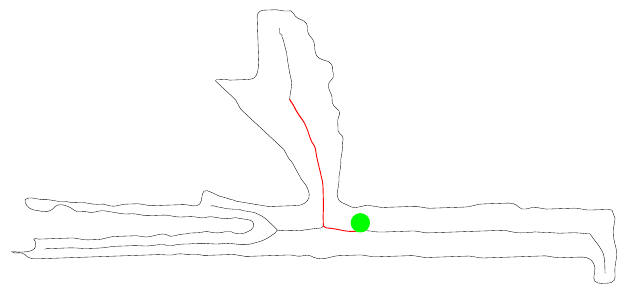} & \includegraphics[width=0.26\textwidth]{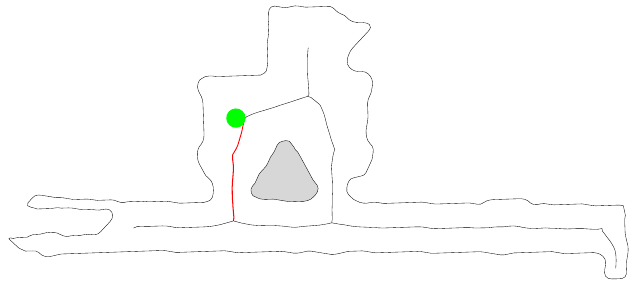} \\[1mm]
			(c) & (d) \\
			\includegraphics[width=0.49\textwidth]{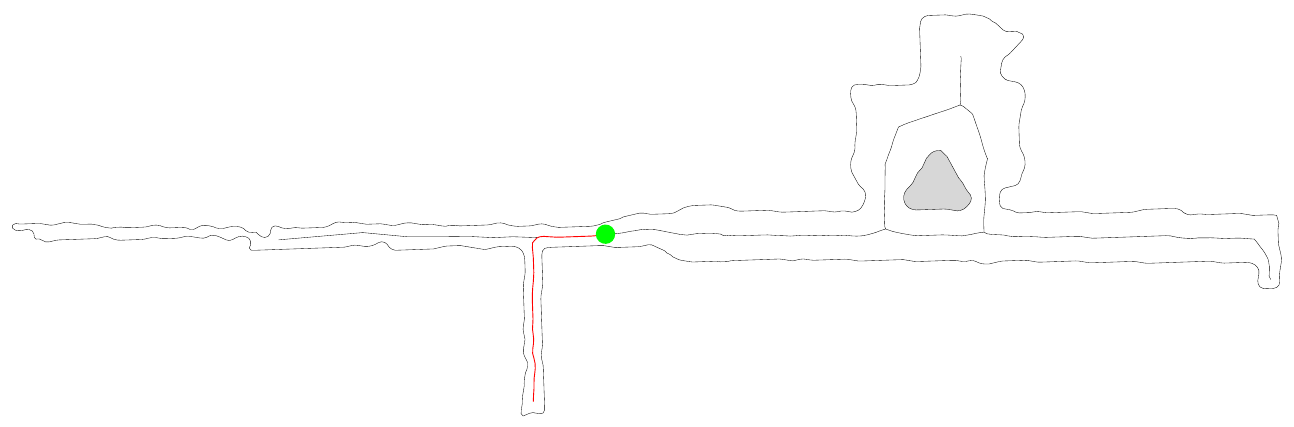} & \includegraphics[width=0.49\textwidth]{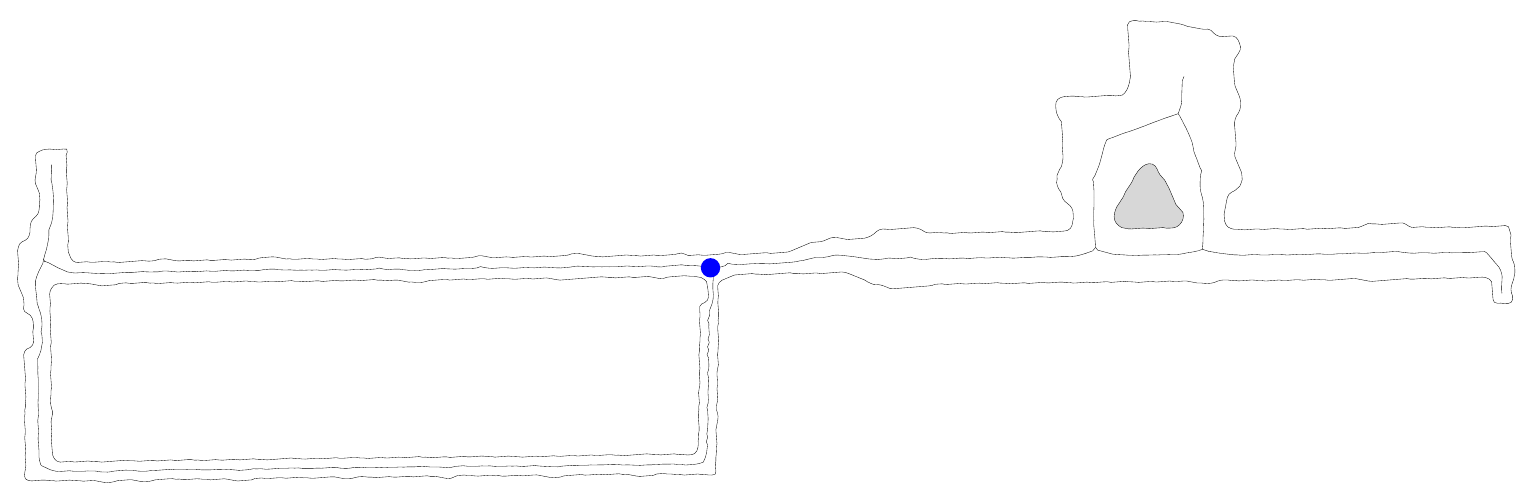}\\ (e) &(f)
			
		\end{tabular}
		
		\caption{Six snapshots from an exploration in the corridors of the McConnell Engineering Building at  McGill University’s buildings. The experiment was conducted using the Turtlebot 2 robot. Similar to Figure  \ref{fig:CaveSlam}, the green disk indicates the position of the robot and the red line of the selected trajectory.  The blue disk indicates successful construction of the skeleton\hyp based map which shows when all the nodes in the topology map are visited.}
		\label{fig:CIM}
		
	\end{figure}

	
	Figure \ref{fig:CIM} presents the proposed algorithm in action using the Turtlebot 2 robot within the corridors of a floor in the McConnell Engineering building of McGill University. We emphasize that the scale of the map changes as the explored environment grows. The robot starts with a very limited view of the environment and the resulting skeleton is a simple curve, the concave part results from the limited field of view of the laser sensor; see Figure  \ref{fig:CIM}a. The robot identifies one side as a dead\hyp end and proceeds down the corridor; see Figure  \ref{fig:CIM}b, until it detects a junction; see Figure  \ref{fig:CIM}c where the robot decides to follow the right side. Figure \ref{fig:CIM}d shows the robot closing a loop, and then continuing down the corridor selecting the left edge, based on proximity; see Figure  \ref{fig:CIM}e. Finally, Figure  \ref{fig:CIM}f presents the completed map of the environment. 
	
	\subsection{Experiments with a Simulator}
	
	\begin{figure}[!htb]
		\centering
		\begin{tabular}{c@{\hskip 2mm}c@{\hskip 2mm}c}
			\includegraphics[width=0.25\textwidth]{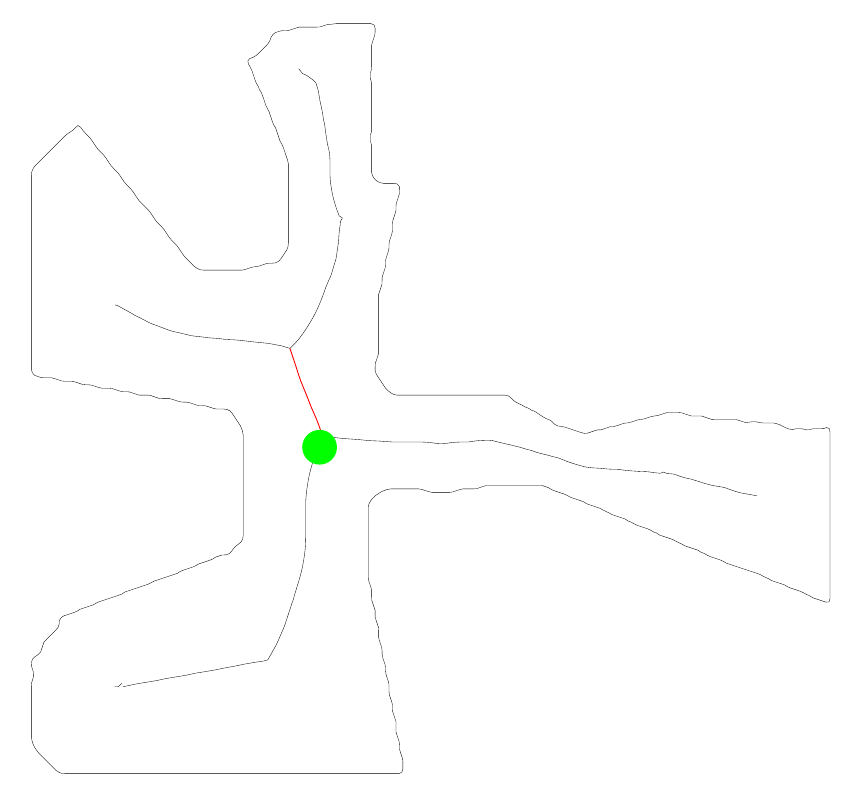} & \includegraphics[width=0.25\textwidth]{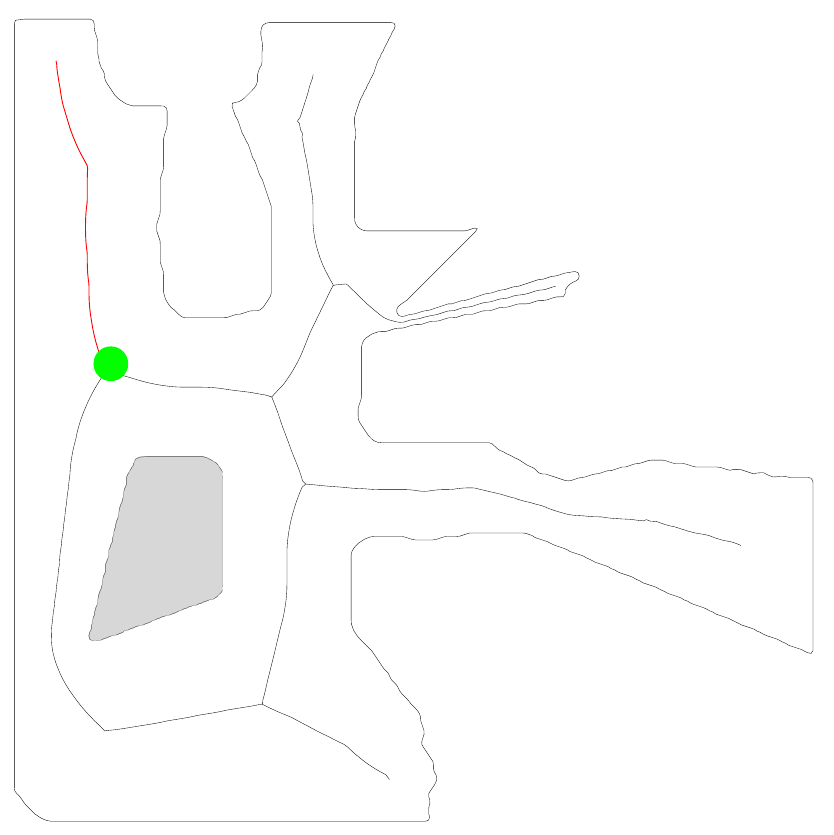}  & \includegraphics[width=0.25\textwidth]{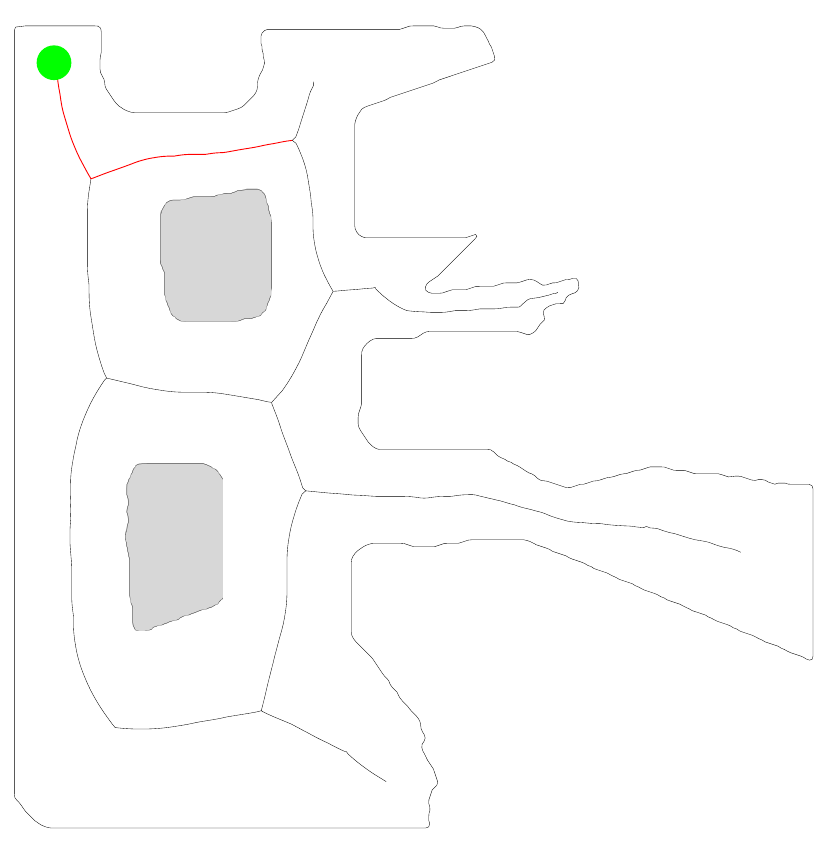} \\
			(a) & (b) & (c) \\[1mm]
			\includegraphics[width=0.25\textwidth]{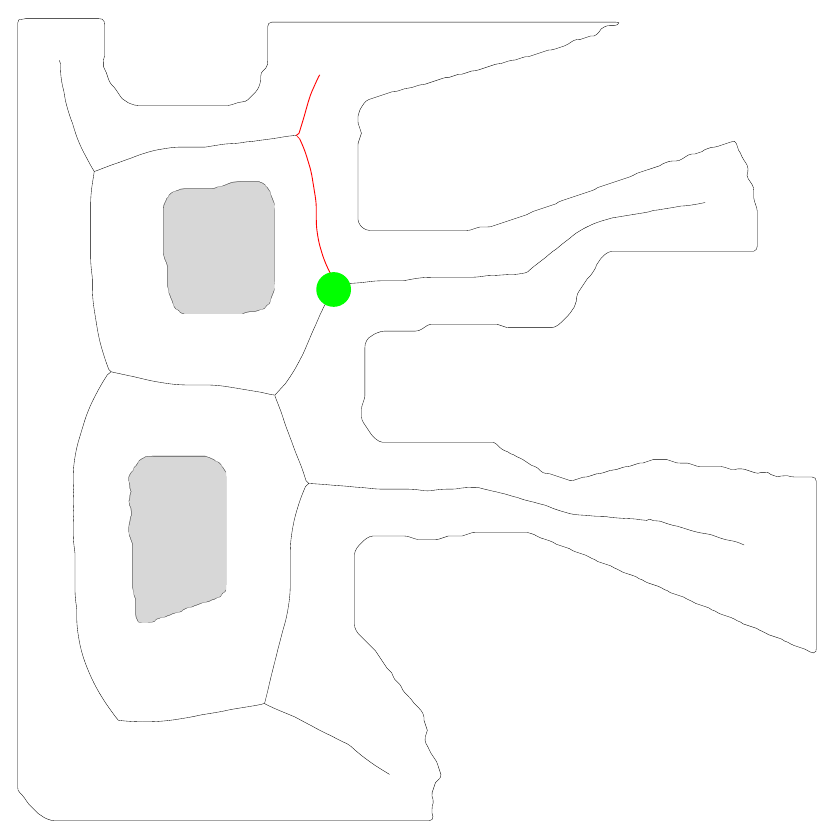} &
			\includegraphics[width=0.25\textwidth]{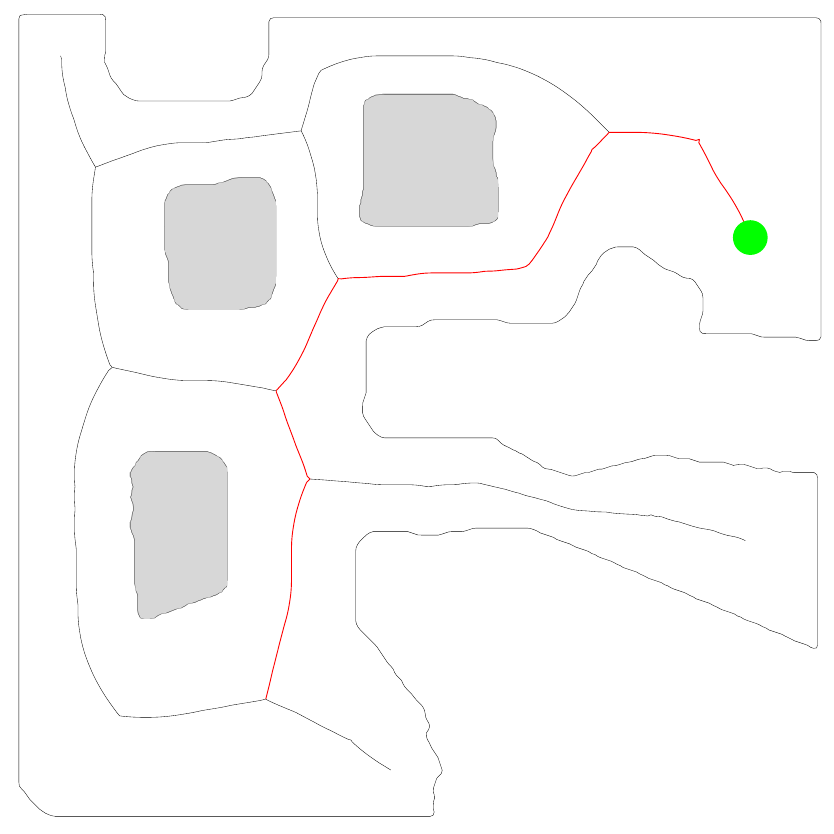} & \includegraphics[width=0.25\textwidth]{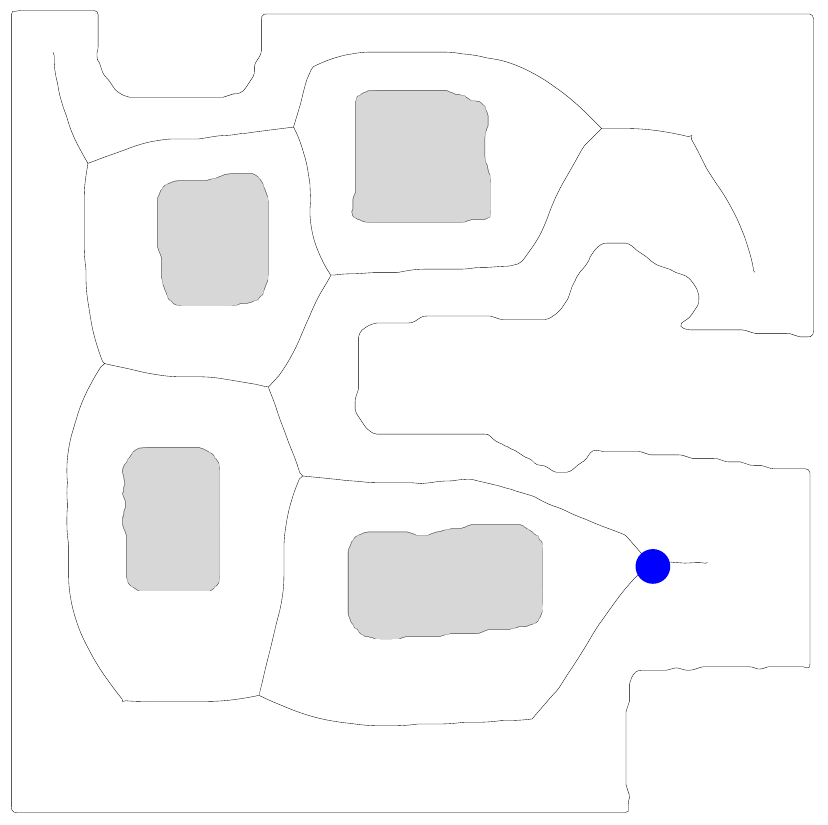} \\
			(d) & (e) & (f) \\
		\end{tabular}    
		\caption{ Six steps of the exploration algorithm, using the Stage cave simulated world, are shown here. At each step, the robot's position, the skeleton of the mapped environment, obstacles, and the future path is shown. The green disk represents the robot, and the red path is where the robot will traverse next. (f) The pose of the robot is drawn in blue to indicate that the robot has now fully explored the map.}
		\label{fig:CaveSlam}
	\end{figure}

	During the simulated experiments, the Stage simulator was used with a different environment, the cave world, as illustrated in Figure \ref{fig:CaveSlam}. The robot started at the middle of the environment, and created a skeleton based on the current information it had (Figure  \ref{fig:CaveSlam}a); after moving to the nearest frontier node, more of the environment became visible and the topological map was updated; see Figure  \ref{fig:CaveSlam}b. The lower left obstacle was not fully mapped, however, enough information was available to produce a loop. In Figure  \ref{fig:CaveSlam}c the robot proceeded to explore the top left corner and then continue the exploration toward the branching nodes at the right side of the environment; see Figure  \ref{fig:CaveSlam}d,e. Finally, the robot finished with a complete topological map of the environment in Figure  \ref{fig:CaveSlam}f.  The robot determines when the navigation through the environment is complete by checking the list of all visited nodes and unexplored nodes on the skeleton.
	

	\section{Topology Matching}
	There are several scenarios that require the matching of two topological maps that are extracted from the same environment in robotics, including map merging, place detection, and map evaluation. One of the ways to evaluate the robustness of an autonomous environment mapping algorithm is to start from different locations in the same environment and then compare the topology maps with each other. If the resulting topology maps are not similar enough, the environment mapping algorithm needs to be revisited in terms of the stability of the representation. Sometimes, an instability in an autonomous algorithm, like ours, can arise from an outside module that has been used in the system. In our case, our environment mapping algorithm can fail if the GMapping module fails. The question of whether our algorithm can recover from these challenging situations then arises. To analyze such failures cases, we evaluate how well areas from one map partially match to areas of another map, i.e., how well the robot can know it has successfully visited an explored area. For these cases, we develop a method that can match two topological maps using the AOF skeleton of their environment, while permitting structural alterations due to perturbations in the sensed boundary.

	To design a matching algorithm for topology maps, we propose using the spectral correspondence algorithm of \cite{lombaert2012focusr}, designed originally for the task of dense vertex correspondence between two surface meshes. In their work, they present an approach to dense vertex-to-vertex correspondence that uses direct matching of features defined on a surface, and then improve it by using spectral correspondence as a regularization. Applications include finding correspondences between meshes undergoing nonrigid transformations and articulated meshes. The algorithm can also be used for precise and accurate correspondence in medical imaging. Representing a structure like a topology map as a graph, a spectral correspondence algorithm tries to characterize that graph via its spectrum, which is the set of eigenvalues and eigenvectors of either the adjacency matrix or the closely related Laplacian matrix. We can think of different ways to represent a matching task as a spectral correspondence process.

	In the following, we will discuss how topology maps are arranged in the form of graph representations and used for spectral correspondence. First, let us assume we have two topology maps and nodes and their connectivity in these maps is represented in the form of a graph. Each skeletal point on the AOF skeleton can be represented by a 3-tuple $(x,y,r)$ where $(x,y)$ represents the skeletal point location and $r$ represents the radius of the corresponding inscribed disk at that particular point. Assume that these two graphs are $G_1 = \{V_1,E_1\}$ and $G_2 = \{V_2, E_2\}$. Note that the correspondence algorithm of \cite{lombaert2012focusr} does not require that $|V_1| = |V_2|$ or  $|E_1| = |E_2|$. The mapping of environments using the FOCUSR algorithm can be described via a four-stage process. What is different here is that rather than choose as an input graph the nodes of a mesh representing the boundary of a 2D or 3D surface, in the current context we shall use the graph based on the AOF skeleton as an input.

	\begin{enumerate}
		\item \textbf{Computing spectra}: In our first configuration setup, a graph is made from the topology map extracted from AOF skeletal points and the connectivity of the nodes is based on whether the medial points are neighboring (in which case we place a value of 1 in the corresponding entry in the adjacency matrix) or not (in which case we place a value of 0 in the adjacency matrix). 
		Having the graph, its adjacency matrix is derived based on the connectivity of nodes in the topology map. In a discrete version of the medial axis as a set of pixels on a grid, two nodes are neighbors if they share either an edge or a corner point. Grady et al. \cite{grady2010discrete} formulated the general Laplacian operator as: 
		\begin{equation}
		\mathcal{L} = G^{-1}(D-W),
		\end{equation}
		where $W$ is a weighted adjacency matrix of the graph with affinity weights (see \cite{grady2010discrete}). 
		We can consider different metrics for the weighted adjacency matrix $W$. In our implementation, we have considered two different metrics for each entry of $W$, $w_{ij}$: a) , $w_{ij} = \frac{1}{dist(i,j)}$, where this value is also multiplied by the connectivity link value between nodes $i$ and $j$ (1 when they are neighbors and 0 when they are not), and b) $w_{ij} = e^{\frac{-dist(i,j)^2}{2*\sigma^2}}$ where like case a) we consider the connectivity link between nodes also. In our first configuration, the distance between two nodes $v_i$ and $v_j$ is computed as: 
		\begin{equation}
		dist(i,j) = \norm{(\mathbf{p}_i,\gamma F_i) - (\mathbf{p}_j,\gamma F_j)}_{2}
		\end{equation}
		where $(\mathbf{p},\gamma F)$ is the concatenation of the 2D coordinate values $\mathbf{p} = (x,y)^T$ with the K feature values $F = (f^{(1)},...,f^{(K)})^T$. $\gamma$ is a $K\times K$ diagonal matrix which contains the $K$ weights controlling the influence of each feature. When there is only one feature, the matrix $\gamma$ reduces to a single number. In our configuration, for every medial point, we consider the radius of the maximal inscribed disk. This feature can play a very important role in matching because thicker (wider) regions get a larger weight in terms of their influence on the matching score. The degree matrix, $D$, is a diagonal matrix, where $D_{ii} = \Sigma_j W_{ij}$, and $G$ can be any meaningful node weighting matrix where nodes with significant features are more heavily favored in the match.  
		
		\item \textbf{Reorder and align spectra}:  When spectra are computed, two situations are possible that make the direct comparison of spectral coordinates challenging. First, computing eigenvectors may generate a sign ambiguity. Second, it is possible that when eigenvectors are being computed for the same value in two maps they might be computed in opposite orders due to the fact that the ordering of the lowest eigenvector may change.  \cite{lombaert2012focusr} suggest mitigating the effects of the flipping problem by favoring three factors: 1) pairs of eigenvectors that are most likely to match based on the similarity between their eigenvalues 2) histograms 3) the spatial distributions of their spectral coordinate value. The process of reordering is sped up by downsampling all eigenvectors. In our experiments, since we are considering points of a 2D medial axis in our configuration (which are far fewer in number than the number of points on a typical 3D surface mesh)  we can consider most of the eigenvectors without worrying about the cost of reordering.

		\item \textbf{Find matches}: After reordering and aligning the spectra, two points that are closest in the embedded representations could be treated as corresponding points in both topology maps. This is achieved by using the Coherent Point
		Drift (CPD) method \cite{myronenko2010point}.

	\end{enumerate}

	\begin{figure}[!h]
		\centering
		\begin{tabular}{@{\hskip 0mm}c@{\hskip 2mm}c@{\hskip 1mm}c}
			\includegraphics[height=0.26\textwidth]{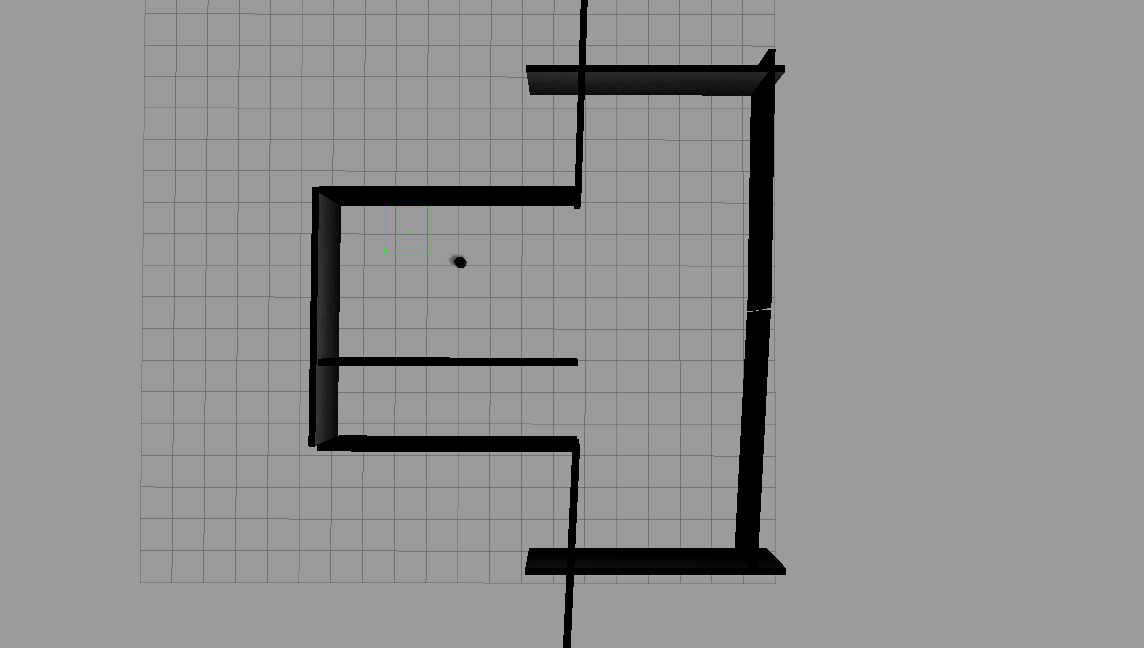} & \includegraphics[height=0.26\textwidth]{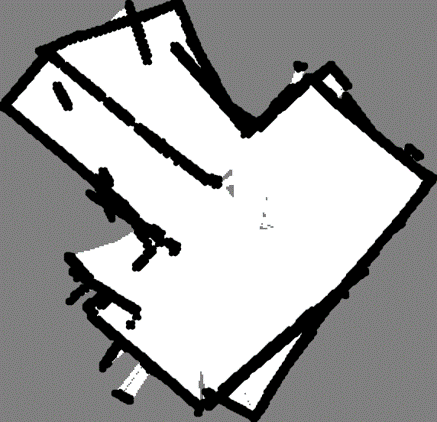} & \includegraphics[height=0.26\textwidth]{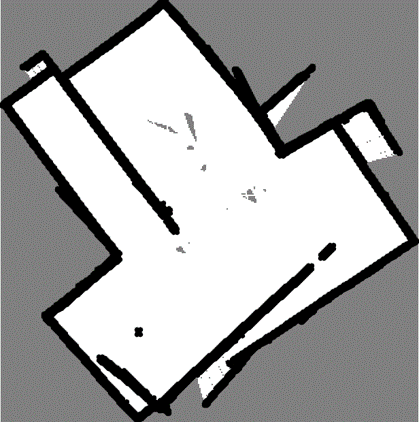}\\
			\textbf{(a)} &  \multicolumn{2}{c}{\textbf{(b)}}
		\end{tabular}
		\caption{\textbf{(a)}: a screenshot of the virtual machine environment with the Gazebo simulator installed. \textbf{(b)}: the two scans of the environment from \textbf{(a)} where for each scan the robot started from a different location. As it can be seen, the left example was deliberately scanned carelessly just to test how robust the algorithm would be in such a circumstance. }
		\label{fig:gazebo_map_occupancy_grid_map}
	\end{figure}
	
	\section{Experiments with Topology Mapping and Matching}
	
	For this project, historically, the initial implementation of the AOF skeleton code was written in C and the stage simulator was used to test the autonomous navigation of the environment. These implementations were the basis of the results published in \cite{rezanejadrobust}. For the topology matching extension, we used a different setup.  First, a virtual machine with ROS Hydro and Gazebo for the Robotics System Toolbox™ was installed on our system. This allowed us to have a ROS Hydro Desktop installation, a Gazebo robot simulator 1.9.6, and some sample Gazebo worlds for a simulated TurtleBot. This allowed us to connect to the virtual machine through its network IP address. This turned out to be extremely useful because we could implement all the other steps in Matlab (see Figure \ref{fig:gazebo_map_occupancy_grid_map} (a)).
	


	To be able to carry out experiments, we ran examples of simulated maps using Gazebo and the Turtlebot robot. For each environment map, we made a launch file with a specific map and a Turtlebot that has a laser scanner range finder installed on it. 

\begin{figure}[!b]
		\centering
		\begin{tabular}{@{\hskip 0mm}c@{\hskip 1mm}c@{\hskip 3mm}c@{\hskip 1mm}c}
			\fbox{\includegraphics[height=0.225\textwidth]{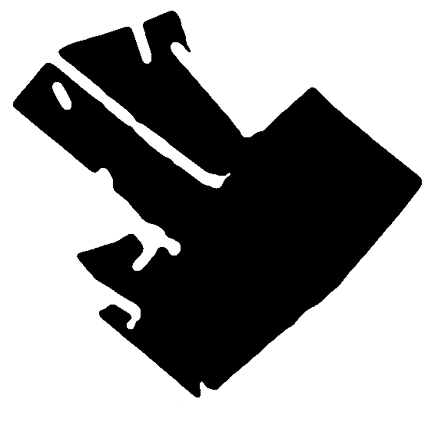}} & \fbox{\includegraphics[height=0.225\textwidth]{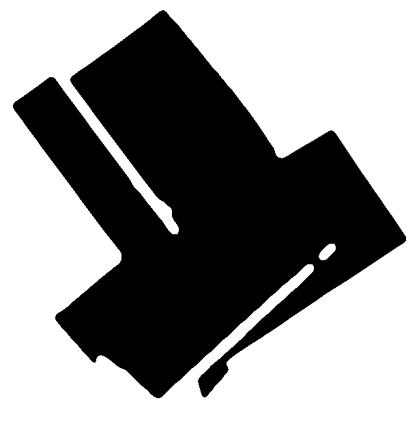}} & 
			\fbox{\includegraphics[height=0.225\textwidth]{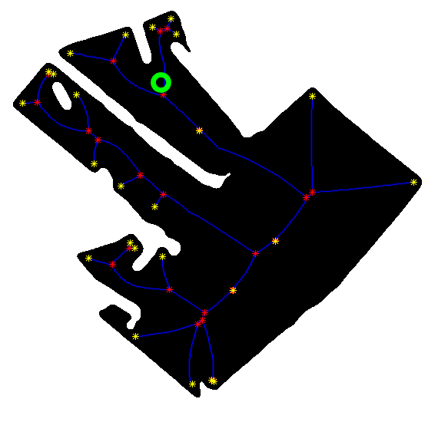}} &
			\fbox{\includegraphics[height=0.225\textwidth]{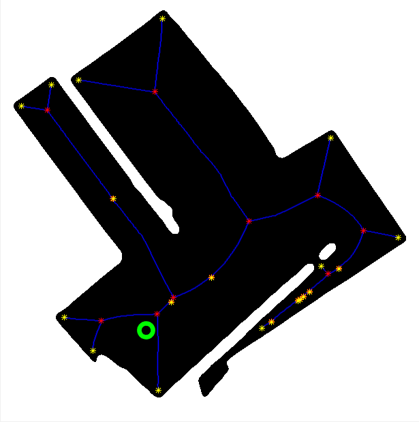}} \\[2mm]
			\multicolumn{2}{c}{\textbf{(a)}} &  \multicolumn{2}{c}{\textbf{(b)}}
		\end{tabular}
		\caption{\textbf{(a)}: these two images show the result of the binarization process for the occupancy grids shown in Figure \ref{fig:gazebo_map_occupancy_grid_map}.
			\textbf{(b)}: these two figures show the result of the skeletonization process on the binary images computed on the left side. The skeletal points are shown in blue here. Branch points are represented by red stars and endpoints are shown by yellow stars. The green disk in each image shows the location where the robot has started the environment mapping. As can be seen, it is not immediately obvious how these two different environment maps could be aligned. To be able to compare these environments, one must work at an appropriate level of abstraction of the skeleton, which is in effect the capability that spectral correspondence provides.}
		\label{fig:binaryImages_skeletonization_result}
	\end{figure}

	Once the GMapping module was working, we carried out an online binarization of images captured from the occupancy grid by writing code in Matlab. We first cropped the occupancy grid map to the area that cells were occupied by values other than the background. Then the image intensities were placed into three categories: 1) scanned pixels 2) obstacles and walls, and 3) background.  To lower the effect of the noise generated by the laser scan and GMapping(e.g. sharp rays), Gaussian blurring with a dynamically chosen smoothing scale was applied (see Figure \ref{fig:gazebo_map_occupancy_grid_map}(b)). The occupancy grid image was then thresholded to give us a binary image (see Figure \ref{fig:binaryImages_skeletonization_result} (a)). 
	
	
	As illustrated in Figure  \ref{fig:binaryImages_skeletonization_result} (b), the skeletonization process produces branches that should be pruned without altering the skeleton's topology. To prune such branches with the goal of topological mapping, we suggest a fairly simple but effective algorithm where the robot explores unseen regions and avoids getting too close to obstacles. Algorithm \ref{alg:pruning} summarizes this process.

	\subsection{FOCUSR Setup}
	To apply the FOCUSR algorithm, we used the MATLAB implementation for matching surfaces introduced in \cite{lombaert2012focusr}. This method matches the meshes of 3D surfaces. We re-coded the package for 2D medial axes by devising 2D medial graphs as follows. We consider $G = (V, E)$, where $V$ represents all medial axis points and $E$ represents their connectivity based on their original connectivity in the skeleton of the environment. Each vertex of $V$ is represented by a quadruple $\mathbf{p}_i = (x_i,y_i,r_i,\theta_i)$, representing the position in $x$-axis, the position in $y$-axis, the radius value at that point (which is the closest distance to the boundary point), and the object angle respectively. Notice that $\theta$ represents the object angle, which is expressed for the unit tangent in the direction of decreasing radius along the medial branch curve. 

	\subsection{Results and Discussion}
\begin{figure}[!htb]
		\centering
		\includegraphics[width=0.75\textwidth]{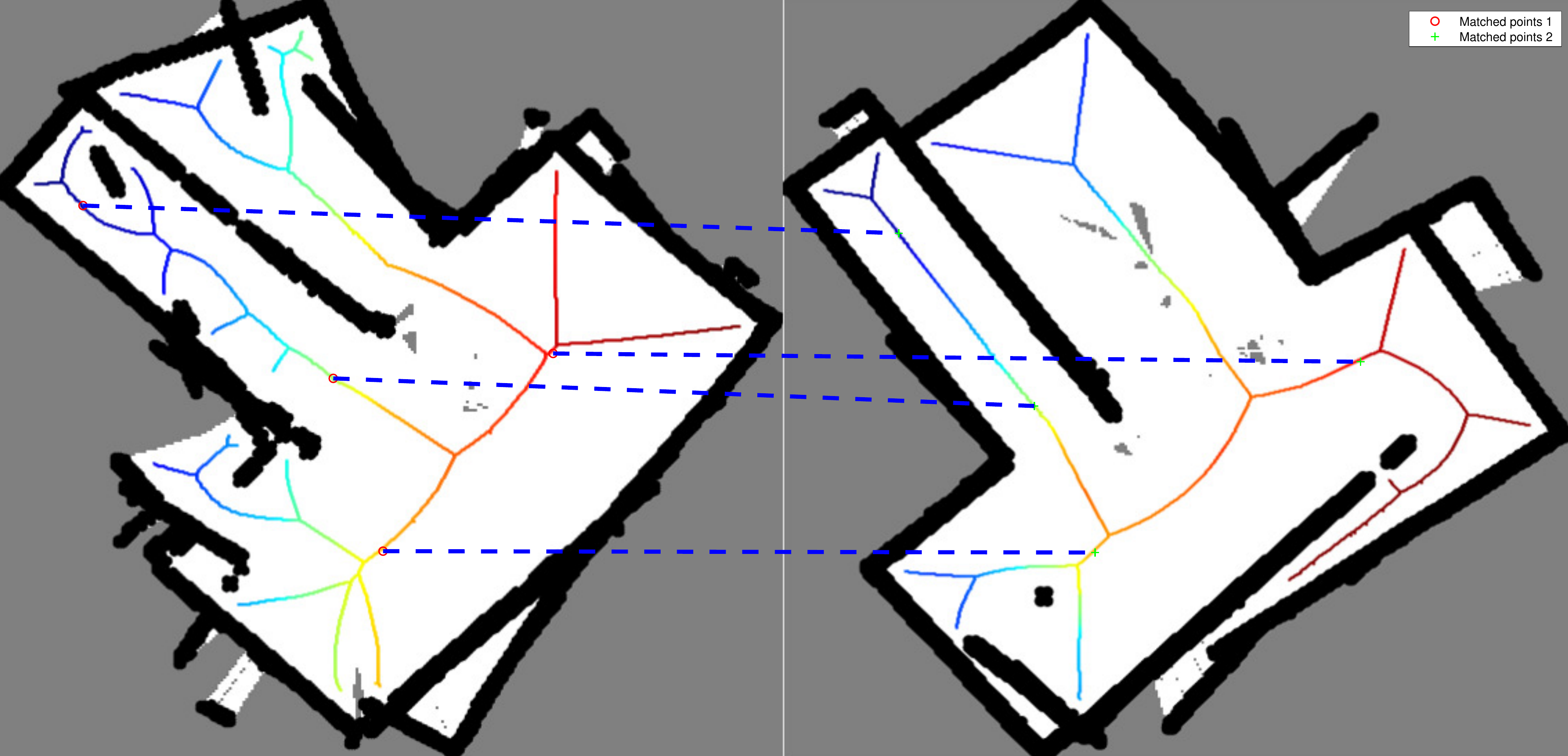}
		
		\caption{ This figure shows the correspondence map between two computed topology graphs, based on their spectral correspondences, with corresponding points shown with similar colors.}
		
		\label{fig:topology_matching}
	\end{figure}

\begin{figure}[!htb]
		\centering
		\begin{tabular}{c@{\hskip3pt}c}
			\includegraphics[height=0.73\textwidth]{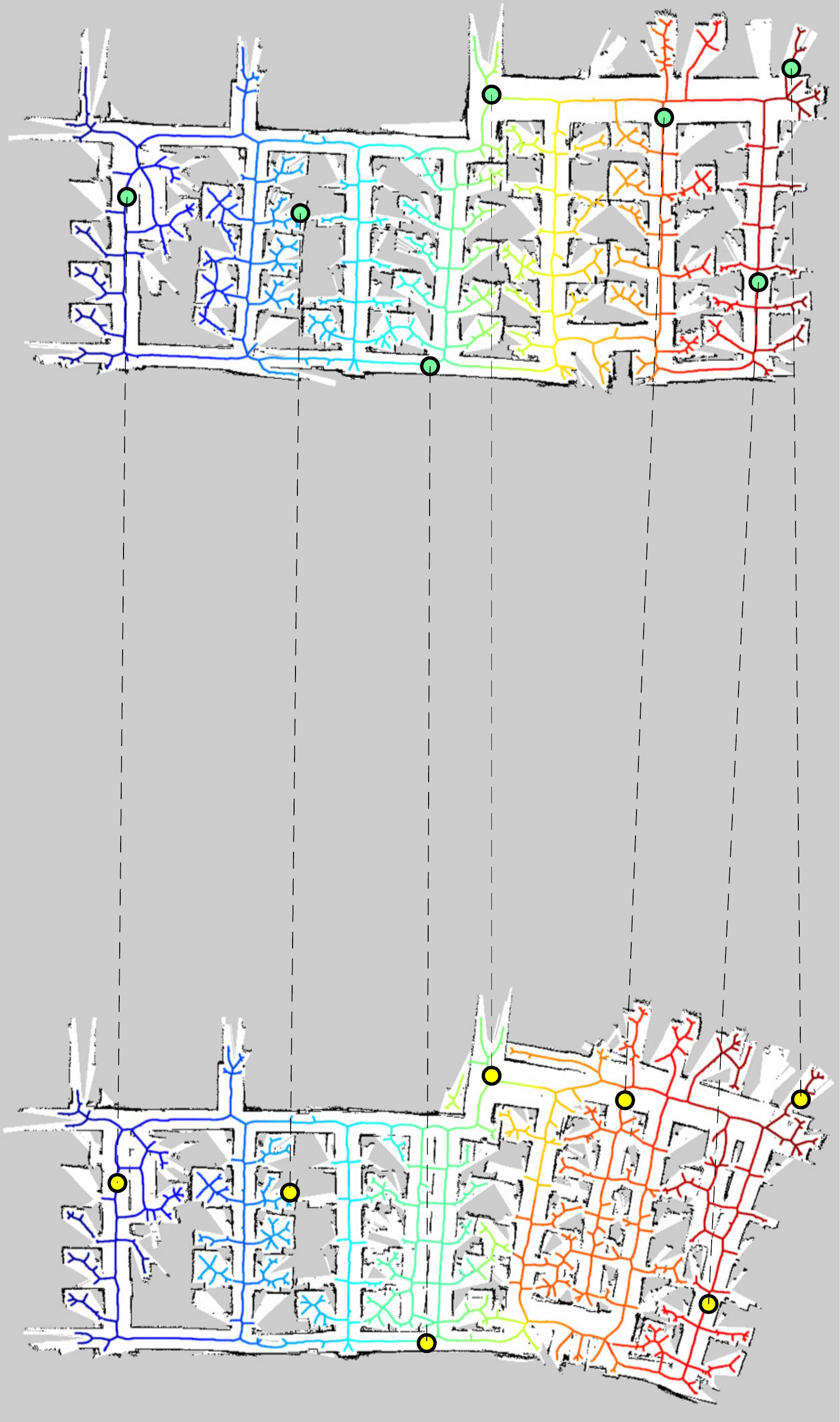}&
			\includegraphics[height=0.73\textwidth]{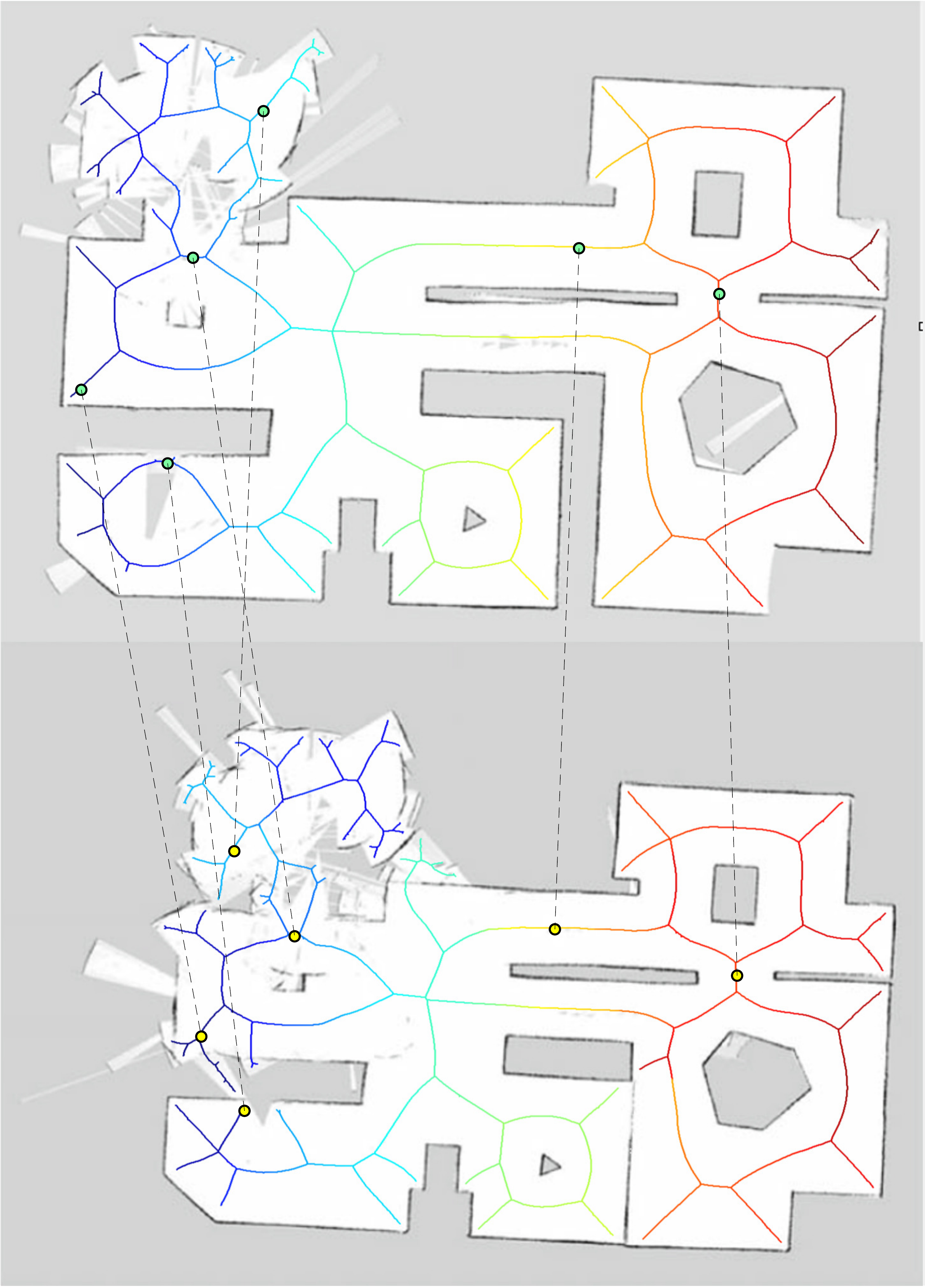}\\
			(a) & (b)\\
		\end{tabular}
		\caption{Same environments mapped differently matched against each other.}
		\label{fig:topology_result2}
	\end{figure}

	We tested the implemented algorithm on two environment maps and we achieved promising results. The visualization shows results that are plausible, i.e., qualitatively similar regions of the maps in terms of spatial layout and local width, seem to align (see Figure \ref{fig:topology_matching}).
	In addition to the previous example, where the robot starts from 2 different locations on the map (Figure \ref{fig:binaryImages_skeletonization_result} (b)), we tried other environments where the GMapping module partially fails in mapping the environment, resulting in occupancy in grid maps that do match exactly to the scenario where the environment is correctly sensed. To see if the maps generated from these situations can still be matched, we present the result of our topology mapping for two additional environments in Figure \ref{fig:topology_result2}. The results show qualitatively plausible matches in terms of the correspondences found between branches of the main (the widest) regions. One of these examples, the one in Figure \ref{fig:topology_result2} (a), is considerably more complex in terms of size and topology than the other examples considered in this article.
	
	Although our algorithm provides some plausible examples of how two environments that are mapped differently can match to each other, we acknowledge that there exists a challenge in evaluating this method quantitatively, and we provide some numbers that put this to perspective. One potential way to quantify this matching is to look at the path similarity between endpoints of the graphs computed from each environment. When correspondences are found between two mapped environments, we can compare each path that connects every two endpoints from one environment to their corresponding ones from another environment. Let us assume we have two explored environments and their medial axes are represented as $G$ and $G'$, and the mapping function that maps the correspondences between them is represented by $\mathcal{T}(\mathbf{p}_i) = \mathbf{p'}_j$, where $\mathbf{p}_i \in G$ and $\mathbf{p}'_j \in G'$. If we let $\mathbf{E} = \{\mathbf{e}_1,\dots,\mathbf{e}_n\}$ and $\mathbf{E}' = \{\mathbf{e}'_1,\dots,\mathbf{e}'_m\}$ represent the endpoints in $G$ and $G'$ respectively, we may then present the following distance measure between these environments as:
	
	\begin{equation}
	\begin{aligned}
	d(G,G') = \frac{\sum\limits_{i=1}^{n}\sum\limits_{j=i+1}^{n} pd(p(\mathbf{e}_i,\mathbf{e}_j),p(\mathcal{T}(\mathbf{e}_i),\mathcal{T}(\mathbf{e}_j)}{n(n-1)} +\\ \frac{\sum\limits_{i=1}^{m}\sum\limits_{j=i+1}^{m} pd(p(\mathbf{e'}_i,\mathbf{e'}_j),p(\mathcal{T}(\mathbf{e'}_i),\mathcal{T}(\mathbf{e'}_j)}{m(m-1)}
	\end{aligned}
	\end{equation}

	where, $p(\mathbf{e}_i,\mathbf{e}_j)$ represents the path between two endpoints $\mathbf{e}_i$ and $\mathbf{e}_j$ in the respective graph (where those nodes belong to) and $pd(\text{path}_1,\text{path}_2)$ represents an  elastic matching between two paths. To compute an elastic matching between two paths, we can use the Optimal Subsequence Bijection method, first presented in \cite{Xu2010-ba}, that accepts two finite
	sequences of end nodes of skeletons and finds the best possible correspondences between them by using a cost function that measures how much those end-to-end points paths are similar to each other.

	
	\section{Discussion}
	\label{sec:conclusions}
	A new methodology for the exploration and mapping of an unknown 2D environment was presented in this article. The algorithm belongs to the family of sensor\hyp based topological maps. It would be useful to carry out a comparison between the use of AOF skeletons, and the use of Voronoi skeletons, as is popular in the literature on environment mapping (\cite{garrido2011path},\cite{choset1997incremental}, and \cite{choset1998sensor}). Carrying out a detailed analysis is beyond the current scope of the work reported here. But it is worth pointing out that the AOF skeletons demonstrate some robustness to a degree of boundary perturbation, an issue that can plague topological representations computed using traditional Voronoi approaches. Utilizing all the recorded data up to the current step results in efficient loop closures and the elimination of the side effects of noise. Experimental results from synthetic as well as live data from an exploring robot demonstrated the efficiency and robustness of the proposed framework. In addition, we proposed a novel spectral correspondence-based matching algorithm between topology maps of environments using AOF skeletons. Experiments show that our algorithm produces promising results, in terms of finding correspondences between similar regions, despite alterations to the graph structures themselves due to simulated sensor noise.

	Future extensions of this work could consider the adaptation of the motion planning technique to deploy on aerial vehicles, such as quadrotors, where the smoothness of the trajectory would be of paramount importance.

	\bibliographystyle{unsrt}  
	\bibliography{references}

\end{document}